%% file: info_neurips_2019.tex
\documentclass{article}

\usepackage[preprint]{neurips_2019}

\usepackage[utf8]{inputenc} 
\usepackage[T1]{fontenc}    
\usepackage{hyperref}       
\usepackage{url}            
\usepackage{booktabs}       
\usepackage{amsfonts}       
\usepackage{nicefrac}       
\usepackage{microtype}      
\usepackage{threeparttable}
\usepackage{amsmath,amsfonts,amsthm} 
\usepackage{multirow}
\newtheorem{lemma}{Lemma}

\usepackage{graphicx}

\title{Utilizing Edge Features in Graph Neural Networks via Variational Information Maximization}

\author{%
	Pengfei Chen$^1$\thanks{Equal contribution},~
	Weiwen Liu$^{1*}$,~
	Chang-Yu Hsieh$^2$,~
	Guangyong Chen$^2$\thanks{Correspondence to: Guangyong Chen <gycchen@tencent.com>.},~
	\textbf{Shengyu Zhang}$^{2}$ \\
	The Chinese University of Hong Kong$^1$,~
	Tencent$^2$ \\
}

\begin{document}

\maketitle

\label{Sec_Abstract}
\input{./sec_Abstract.tex}

\section{Introduction}
\label{Sec_Introduction}
\input{./sec_Introv2.tex}

\section{Related works}
\label{Sec_Related}
\input{./sec_Related.tex}

\section{Our method}
\label{Sec_Method}
\input{./sec_Method.tex}

\section{Experiments}
\label{Sec_Experiments}
\input{./sec_Experiments.tex}

\section{Conclusions}
\label{Sec_Conclusions}
\input{./sec_Conclusions.tex}


\bibliography{neurips_2019}
\bibliographystyle{apalike}

\medskip

\small

\end{document}

%% file: sec_Abstract.tex
\begin{abstract}
  Graph Neural Networks (GNNs) achieve an impressive performance on structured graphs by recursively updating the representation vector of each node based on its neighbors, during which parameterized transformation matrices should be learned for the node feature updating. However, existing propagation schemes are far from being optimal since they do not fully utilize the relational information between nodes. We propose the information maximizing graph neural networks (IGNN), which maximizes the mutual information between edge states and transform parameters. We reformulate the mutual information as a differentiable objective via a variational approach. We compare our model against several recent variants of GNNs and show that our model achieves the state-of-the-art performance on multiple tasks including quantum chemistry regression on QM9 dataset, generalization capability from QM9 to larger molecular graphs, and prediction of molecular bioactivities relevant for drug discovery. The IGNN model is based on an elegant and fundamental idea in information theory as explained in the main text, and it could be easily generalized beyond the contexts of molecular graphs considered in this work. To encourage more future work in this area, all datasets and codes used in this paper will be released for public access.
  
\end{abstract}

%% file: sec_Introv2.tex
Many real-world datasets naturally come in the form of graphs, such as social networks, maps, knowledge graphs, molecules, protein-protein interactions etc., all of which consist of a number of nodes and edges equipped with their inherent features. It has been a fundamental task to predict some properties of a single node or a whole graph, such as predicting the community belonging of a user in a social network and predicting the binding score in a virtual-docking task for a drug discovery. Recently, impressive performance has been achieved by Graph Neural Networks (GNNs)~\cite{gori2005new,scarselli2009graph} and their various variants \cite{duvenaud2015convolutional,defferrard2016convolutional,li2016gated,kipf2017semi,hamilton2017inductive,gilmer2017neural,velivckovic2018graph,schlichtkrull2018modeling,liao2019lanczosnet}. Compared with previous efforts, such as node2vec \cite{grover2016node2vec}, GNNs learn the state of a node by recursively aggregating the states of its neighbor nodes, which incorporates the graph structure with node features.  Intuitively, edge features also play crucial roles in determining the classification or regression results. For example, chemical bonds in a molecule are important when we want to predict molecular properties.  Unfortunately, not much efforts have been devoted to building models that better utilize the edge attributes on a graph until now. Our proposed model utilizes edge features better than all existing models in the literature.

%

The expressive power of GNNs largely depends on how the message is passed between nodes, namely the transformation of node states during propagation. A widely adopted transformation is multiplying node states with a parameterized matrix before aggregation. Despite tremendous success of GNNs on graph structured data, message passing schemes in existing models are far from being optimal since they do not fully utilize the edge information between nodes. For example, many early variants of GNNs such as Graph Convolutional Networks (GCN) \cite{kipf2017semi}, Chebyshev Networks (ChebyNet) \cite{defferrard2016convolutional} and Graph Attention Network (GAT) \cite{velivckovic2018graph} do not even support categorized edge types. In these models, states of all neighbor nodes are transformed by multiplying the same transform matrix, which is simply a trainable parameterized matrix or parameterized by a fully connected neural network layer. 

Augmenting GNNs' representability by incorporating edge features has been demonstrated in some previous works. To consider edge features in highly multi-relational data, different transform matrices are introduced for different edge types. This strategy is adopted in several variants such as Relational Graph Convolutional Networks (RGCN) \cite{schlichtkrull2018modeling}, Gated Graph Neural Networks (GGNN) \cite{li2016gated} and Lanczos Networks (LanczosNet) \cite{liao2019lanczosnet}, all of which, however, take into account discrete edge types and cannot handle continuous edges states. To handle general feature vectors of edges, Message Passing Neural Networks (MPNN) \cite{gilmer2017neural} introduces an edge network that takes edge features as input and outputs transform matrices, then the edge-specific transform matrices are used to transform states of neighbor nodes.  

Thanks to the edge network, MPNN \cite{gilmer2017neural} exhibits high flexibility in handling various kinds of edge states. For example, in molecular graphs, discrete bond type is encoded as one-hot vector while continuous distance between atoms is encoded as a scalar. These distinct edge features can be easily merged as one edge feature vector. However, simply feeding the feature vectors of edges to a neural network and obtaining large transform matrices seems to be primitive, and such a scheme still does not ensure the features of edges are fully utilized in the propagation of GNNs. While in many real-world applications such as inferring properties of molecules, edge features are of great importance. For example, in molecular graphs, distances between atoms are closely related to the degree of bond strength that could not be inferred from the topological structures of a molecular graph alone.  Therefore, a GNN model that could better utilize the edge features should deliver better performance on molecular property prediction. 

In this paper, we attempt to preserve as much edge information in the propagation of GNNs as possible. To achieve this, we follow the widely-accepted message passing framework (transformation and aggregation) and propose to maximize the mutual information between edge states and transform parameters of node states, namely Information Maximizing Graph Neural Networks (IGNN). However, computing the mutual information requires access to the posterior distribution of edge states conditional on transform parameters, which is intractable in practice. To address this issue, we adopt a variational approach \cite{agakov2004algorithm} to reformulate the mutual information as an easily accessible differentiable objective, which is applied as a regularization term during training. In this way, IGNN explicitly reduces information loss of edge features during propagation, hence exhibits superiority to existing GNNs on highly multi-relational graphs. Our main contributions can be summarized as follows:

\begin{itemize}
	\item We propose a variational mutual information maximizing approach for GNNs, which maximizes the mutual information between edge states and transform parameters of node states. The approach is proposed with a theoretical information lower bound.
	\item We experimentally demonstrate that our IGNN achieves the state-of-the-art performance on the benchmark dataset QM9 \cite{ramakrishnan2014quantum} when compared with extensive baselines. Our results explicitly manifest that edge features are crucial for predicting molecule properties. 
	\item We empirically demonstrate that generalization on molecules with larger size still remains as a challenging task, and our IGNN achieves impressive performance in comparison to most baselines.
	\item Finally, we demonstrate the potential usefulness of IGNNs on drug discovery, i.e., predicting the molecular bioactivity with respect to the  lung cancer related target ERBB1.  IGNN consistently achieves the state-of-the-art performance in this task.
	
\end{itemize}

%% file: sec_Related.tex
\paragraph{Preliminaries.} Let $G=(\mathcal{V},\mathcal{E})$ be a graph with node feature vectors $x_v\in\mathbb{R}^{d}$ for node $v\in \mathcal{V}$ and edge feature vectors $e_{vw}\in\mathcal{E}$ for the edge connecting node $v$ and $w$. In the propagation of GNNs, the state of each node are updated recursively using the states of neighbor nodes. Let $\mathcal{N}_v$ be the set of neighbor nodes of $v$ and $h_v^{(l)}\in\mathbb{R}^{d_l}$ be the hidden state of $v$ at $l$-th layer, where $d_l$ is the dimension of the hidden layer. For simplicity of notation, we use a single $d$ to denote the dimension such that  $h_v^{(l)}\in\mathbb{R}^{d}$. We also have $h_v^{(0)}=x_v$ at the input layer. In node classification and regression tasks, we usually focus on a single graph $G$ and each node $v\in V$ is associated with a label $y_v$. In graph classification and regression tasks, we have a collection of graphs $\{G_1,...,G_N\}$ and their labels $\{y_1,...,y_N\}$, one for each graph.
\subsection{Propagation in Graph Neural Networks}

In this part, we briefly review some of the most common GNNs with focus on handling edge features during the propagation, which is closely related to our method.

\paragraph{Single-relational modeling.} Many variants such as GCN \cite{kipf2017semi}, GAT \cite{velivckovic2018graph}, ChebyNet \cite{defferrard2016convolutional}, GraphSAGE \cite{hamilton2017inductive} focus on learning node state. These models can assign weight to neighbors, but they do not discuss categorized edge types. For example, in the propagation of GCN, the typical neighborhood aggregation scheme is written as
\begin{equation}
\label{Eg_gcn}
h_v^{(l+1)}=\sigma\left(\sum_{w\in \mathcal{N}_v}\dfrac{1}{c_{vw}}W_{1}^{(l)}h_w^{(l)} + W_{0}^{(l)}h_v^{(l)}\right),
\end{equation}
where $\sigma$ denotes an activation function such as ReLU, $c_{vw}$ is a normalization constant such as $c_{vw}=\sqrt{deg(v)deg(u)}$ and $deg(v)$ is the degree of node $v$ in $G$. Without considering different edge types, states of all neighboring nodes are multiplied by the same transform matrix $W_{1}^{(l)}$, which is simply a trainable parameterized matrix or parameterized by a fully connected neural network layer. Sometimes the self-connection are also treated in the same way so that we have $W_{0}^{(l)}=W_{1}^{(l)}=W^{(l)}$. Following the same framework, skip connections or trainable neighboring weighting schemes are introduced in other variants such as GAT, ChebyNet and GraphSAGE, yet the handle of highly multi-relational graph are not discussed in this line of work.

\paragraph{Multi-relational modeling.} A simple strategy to handle multi-relational graph is assigning each edge type with a specific transform matrix. This strategy is adopted in several variants such as RGCN \cite{schlichtkrull2018modeling}, GGNN \cite{li2016gated} and LanczosNet \cite{liao2019lanczosnet}. For example, RGCN updates node states according to the following scheme
\begin{equation}
\label{Eg_rgcn}
h_v^{(l+1)}=\sigma\left(\sum_{r\in \mathcal{R}}\sum_{w\in \mathcal{N}_v^r}\dfrac{1}{c_{vw,r}}W_{r}^{(l)}h_w^{(l)} + W_{0}^{(l)}h_v^{(l)}\right),
\end{equation}
where $\mathcal{N}_v^r$ is the collection of neighboring nodes of $v$ with relation $r\in\mathcal{R}$ and $c_{vw,r}$ is a normalization constant such as $c_{vw}=\lvert\mathcal{N}_v^r\rvert$. Such a scheme can not handle continuous edges states since assigning different transform matrices $W_{r}^{(l)}$ for all edge types is impossible in continuous case, where $\lvert\mathcal{R}\rvert$ is $\infty$. Although continuous edge features may be embedded in the normalization constant $c$, this does not really solve the problem since in this way, we are actually assigning a scale factor to $W_{r}^{(l)}$, which limits the diversity of the mapping from edge features to $W$. GGNN introduces Gated Recurrent Unit (GRU) \cite{cho2014properties} to treat states of each node across layers as a sequence, yet does not focus on the improvement of edge expressibility. Instead, GGNN simplifies the edge expressibility by sharing the transform matrix across layers, such that $W_{r}$ is related to the relation but invariant across layers. LanczosNet takes into account multi-scale connections. A novelty of LanczosNet is using the Lanczos algorithm \cite{lanczos1950iteration}
to construct low rank approximations of the graph Laplacian, enabling efficient exploitation of multi-scale information in graphs. LanczosNet adopts the same strategy to model multi-relational graph as RGCN, hence still can not handle complex edge features.

\paragraph{Complex-relational modeling.} The relation in graph can be quite complex, such as for molecules, we have discrete bond types (single,
double, triple, or aromatic) and continuous distance between atoms. These edge features can be expressed as general feature vectors $e_{vw}$. To handle complex edge feature vectors, MPNN \cite{gilmer2017neural} introduces an edge network which takes as input feature vectors of edges and output transform matrices. A single edge network is shared throughout the MPNN model. The propagation is formalized as
\begin{equation}
\label{Eg_mpnn}
\begin{aligned}
&m_v^{(l+1)}=\sigma\left(\sum_{w\in \mathcal{N}_v}f(e_{vw})h_w^{(l)} + W_{0}^{(l)}h_v^{(l)}\right),\\
&h_v^{(l+1)}={\mathrm{GRU}}(h_v^{(l)},m_v^{(l+1)}),
\end{aligned}
\end{equation}
where $f: e\rightarrow W$ denotes the edge network expressed as a neural network. In this way, $e_{vw}$, taking the form of a general feature vector, is associated with a transform matrix $W_{vw}=f(e_{vw})$. In MPNN, the update of node states follows the method proposed by GGNN \cite{li2016gated}, where states of each node across layers are treated as a sequence and updated by GRU\cite{cho2014properties}.

\paragraph{Summary.} 

Due to the introduction of an edge network, MPNN \cite{gilmer2017neural} does outperform other methods by a large margin on QM9 \cite{ramakrishnan2014quantum}. However, simply feeding the feature vectors of edges to a neural network and obtaining large transform matrices seems to be primitive, and such a scheme still does not ensure the features of edge are fully utilized in the propagation of GNNs. Therefore, we are motivated to optimize the propagation by maximizing the mutual information between edge features and transform matrices. The detailed approach will be discussed in Sec.~\ref{Sec_Method}.

\subsection{Readout functions.}
After several propagations, we obtain final states of all nodes, which are suitable for node classification and regression. For tasks such as graph classification and regression, we can apply a readout function \cite{ying2018hierarchical,vinyals2015order} such that
\begin{equation}
\label{Eq_readout}
\hat{y}=R(\{h_v^{L},|v\in G\}),
\end{equation}
where $h_v^{L}$ is the state of $v$ at the end of propagation, $R$ is the readout function that takes as input a set of node states and outputs a graph-level representation $\hat{y}$. For example, we can use the set2set model \cite{vinyals2015order}, or simply sum up the final node states \cite{liao2019lanczosnet,xu2019powerful}. 

%% file: sec_Method.tex
Mutual information inspired objective functions have long been adopted in unsupervised learning \cite{bridle1992unsupervised,barber2006kernelized,hjelm2018learning,velivckovic2018deep}, semi-supervised classification \cite{krause2010discriminative} and generative adversarial networks \cite{chen2016infogan}. Notably, Deep Graph Infomax (DGI) \cite{velivckovic2018deep} proposes to learn node-wise representations in an unsupervised manner by maximizing the mutual information between node representations and corresponding high-level summaries of graphs. Our objective and method are absolutely different from DGI.

Our method targets at preserving edge information in the propagation of GNNs, which is important in many real-world graphs such as molecules - apart from node (atom) features, attributes of edges (bonds) are quite important for predicting properties of graphs. To achieve the goal of fully utilizing edge features, we propose to maximize the mutual information between the edge feature vector $e$ and the transform matrix $W$, where $W=f(e)$ is given by $f:e\rightarrow W$ and $W$ is used to transform the node states during propagation, as introduced in Sec.~\ref{Sec_Related}. To show the wide applicability of our method, here we do not have any assumption on the specific form of $f$. Depending on the implementation, $f$ can be a neural network that takes as input a general form of edge feature vector, such as in MPNN \cite{gilmer2017neural}; or simply a discrete mapping such that $f$ maps any relation $r\in\mathcal{R}$ to a transform matrix $W_r$, such as in GGNN \cite{li2016gated}. Our method can be easily generalized to any GNNs that require a edge-related transform matrix $W$.

\subsection{Variational mutual information}
\label{Sec_Method1}
Computing the mutual information $I(e;W)$ requires access to the posterior $p(e|W)$, which is intractable in practice. Thus, we adopt a variational approach \cite{agakov2004algorithm} to reformulate $I(e;W)$ as a differentiable objective.

In our setting, the prior $p(W|e)$ is simply given by $f$ such that 
\begin{equation}
\label{Eq_prior}
p(W|e)=\delta(W-f(e)),
\end{equation}
where $\delta(\cdot)$ is the Dirac delta function.
The posterior $p(e|W)$ is intractable, so we define a variational distribution $q(e|W)$, which can be obtained by defining a neural network $g:W\rightarrow e$. Specifically, $q(e|W)$ substitutes to some distribution (such as Gaussian distribution) with parameter $g(W)$. In this way, $f$ and $g$ are similar to the probabilistic encoder and decoder in the Variational Auto-Encoder (VAE) \cite{kingma2013auto}. Then we can approximate $I(e;W)$ with a differentiable objective $L_I(f,g;e)$ as follows.
\begin{lemma}
\label{Lemma_1}
Let $e$ be the edge feature vector, $W$ be the transform matrix with conditional distribution $p(W|e)$ specified by the probabilistic encoder $f$ as shown in Eq.~(\ref{Eq_prior}) and $q(e|W)$ be the variational distribution specified by the probabilistic decoder $g$, then we have
\begin{equation}
\label{Eq_var_info}
I(e;W)\geq H(e)+\mathbb{E}_{e\sim p(e)}[\mathcal{L}_{I}(f,g;e)],
\end{equation}
where $\mathcal{L}_{I}(f,g;e)=\log q(e|f(e))$ and $H(\cdot)$ denotes the entropy.
\begin{proof} Let $D_{KL}(\cdot\parallel \cdot)$ denote the KL-divergence, which should be nonnegative, then we have  
\begin{equation}
\nonumber
\begin{aligned}
I(e;W)&=H(e)-H(e|W)\\
&=H(e)+\mathbb{E}_{W\sim p(W)}[\mathbb{E}_{e\sim p(e|W)}[\log p(e|W)]]\\
&=H(e)+\mathbb{E}_{W\sim p(W)}[\mathbb{E}_{e\sim p(e|W)}[\log p(e|W)-\log q(e|W)+\log q(e|W)]]\\
&=H(e)+\mathbb{E}_{W\sim p(W)}[D_{KL}(p(e|W)\parallel q(e|W))+\mathbb{E}_{e\sim p(e|W)}[\log q(e|W)]]\\
&\geq H(e)+\mathbb{E}_{W\sim p(W)}[\mathbb{E}_{e\sim p(e|W)}[\log q(e|W)]]\\
&=H(e)+\mathbb{E}_{e\sim p(e),W\sim p(W|e)}[\log q(e|W)]\\
&\overset{(a)}{=}H(e)+\mathbb{E}_{e\sim p(e)}[\log q(e|f(e))]\\
\end{aligned}
\end{equation}
where the final equality $(a)$ follows from Eq.~(\ref{Eq_prior}).	
\end{proof}
\end{lemma}  
According to Lemma~\ref{Lemma_1}, we propose to maximize the variational lower bound for the mutual information $I(e;W)$. The bound becomes tight when the variational distribution $q(e|W)$ approaches the true posterior $p(e|W)$. $H(e)$ is a constant since the distribution of edge feature vector $e$ is fixed for given graphs, hence we can equivalently maximize $\mathcal{L}_{I}(f,g;e)$. To implement it as a differentiable objective using the neural networks $f$ and $g$, we should choose the prior distribution of the probabilistic decoder $g$. Here we give an example of the widely used Gaussian prior such that
\begin{equation}
q(e|W)=\mathcal{N}(e;g(W),\sigma^2I).
\end{equation}
Then we have 
\begin{equation}
\label{Eq_LI}
\begin{aligned}
\mathcal{L}_{I}(f,g;e)=\log q(e|f(e))=\log \mathcal{N}(e;g(f(e)),\sigma^2I)=-\lambda \lVert e-g(f(e))\rVert_2^2
\end{aligned}
\end{equation}
where $\lambda>0$ is a constant determined by $\sigma$ and the dimension of $e$, taken as a tunable parameter. 

\subsection{Information Maximizing Graph Neural Networks}
In this section, we show how to implement the variational mutual information objective in GNN, and derive our Information Maximizing Graph Neural Networks (IGNN). As a concrete example, the propagation of our model follows the formulation in Eq.~(\ref{Eg_mpnn}), where $f:e\rightarrow W$ is expressed as a neural network, shared across all layers. According to the theoretical analysis in Sec.~\ref{Sec_Method1} , we introduce a neural network $g:W\rightarrow e$. The $f$ and $g$ can be implemented with multi-layer perceptrons (MLPs).

For graph regression or classification tasks, the propagation and readout yield a prediction $\hat{y}$ for each graph $G$, which has label $y$. Without variational mutual information maximization, we denote the traditional loss as $\mathcal{L}_{0}(\hat{y},y;G)$. Common choice of $\mathcal{L}_{0}$ includes Mean Square Error (MSE), Mean Absolute Error (MAE) and Cross Entropy (CE). In IGNN, we should maximize $\mathcal{L}_{I}(f,g;e)$ and minimize $\mathcal{L}_{0}(\hat{y},y;G)$ simultaneously. For a graph $G=(\mathcal{V},\mathcal{E})$, the objective function of IGNN is given by
\begin{equation}
\label{Eq_IGNN_Graph}
\mathcal{L}(\hat{y},y;G)=\mathcal{L}_{0}(\hat{y},y;G)-\mathbb{E}_{e\in \mathcal{E}}[\mathcal{L}_{I}(f,g;e)],
\end{equation}
where $\mathbb{E}_{e\in \mathcal{E}}[\cdot]$ denotes taking the mean over all edges in $G=(\mathcal{V},\mathcal{E})$. We minimize the loss $\mathcal{L}(\hat{y},y;G)$ during training. When IGNN is trained using mini-batches, $\mathcal{L}_{0}(\hat{y},y;G)$ is averaged over all graphs in the batch while $\mathcal{L}_{I}(f,g;e)$ is averaged over all edges of all graphs in the batch.

Similarly, for node regression or classification tasks, the propagation of IGNN directly yields node-level representations $\hat{y}$ for each node $v\in \mathcal{V}$ and the objective function of IGNN can be written as
\begin{equation}
\label{Eq_IGNN_Node}
\mathcal{L}(\hat{y},y;G)=\mathbb{E}_{v\in \mathcal{V}}[\mathcal{L}_{0}(\hat{y},y;v)]-\mathbb{E}_{e\in \mathcal{E}}[\mathcal{L}_{I}(f,g;e)],
\end{equation}
where $\mathcal{L}_{0}(\hat{y},y;v)$ is a traditional loss associated with node $v$ and $\mathbb{E}_{v\in \mathcal{V}}[\cdot]$ denotes taking the mean over all nodes in $G=(\mathcal{V},\mathcal{E})$.

%% file: sec_Experiments.tex
Our proposed model is evaluated and compared with existing baselines under two scenarios --- {\bf predicting quantum-mechanism properties} and {\bf drug discovery}. Feature engineering of nodes and edges exactly follows the setting in \cite{gilmer2017neural} for both scenarios. 
We compare against nine state-of-the-art baselines, \emph{i.e.,} Graph Convolutional Networks (GCN)~\cite{kipf2017semi}, Relational Graph Convolutional Networks (RGCN)~\cite{schlichtkrull2018modeling}, Chebyshev networks (ChebyNet)~\cite{defferrard2016convolutional}, Graph Attention Network (GAT)~\cite{velivckovic2018graph}, Gated Graph Neural Networks (GGNN)~\cite{li2016gated}, Lanczos Networks (LanczosNet)~\cite{liao2019lanczosnet}, Graph Isomorphism Network (GIN)~\cite{xu2019powerful}, simplified Message Passing Neural Networks (sMPNN) with only categorical edge features, and Message Passing Neural Networks (MPNN)~\cite{gilmer2017neural} with both categorical and continuous edge features. 

For the fairness of our experiments, we use Adam~\cite{kingma2014adam} as the optimization method, and the dimension of hidden layers is 64 for all the experiments. The same random seed is shared for all models. For LanczosNet and GGNN, the implementation of the readout function follows the original paper. While for all other models, we use the same readout, set2set \cite{vinyals2015order}, which has been demonstrated to work well in \cite{gilmer2017neural}.  The tunable parameter $\lambda$ is chosen as $1$ naively for  all experiments. 

\subsection{Quantum chemistry}\label{subsect:quantum_chemistry}

Quantum property regressions are computed on benchmark dataset QM9  \cite{ramakrishnan2014quantum}, which contains 134K molecules comprised of up to 9 heavy atoms excluding the Hydrogen H. All models are trained to predict the 12 target properties, including harmonic frequencies, dipole moments, polarizabilities, electronic energies, enthalpy, and free energies of atomization. Each target property is normalized to zero mean and unit variance. We randomly choose 10,000 molecules for validation, 10,000 molecules for testing, and keep the rest for training. The validation set is used for early stopping and model selection. We use mean square error (MSE) loss to train the models over 300 epochs, and the performances are measured by mean absolute error (MAE).


\begin{table}[tp]
\centering
\scriptsize
\begin{threeparttable}
\caption{Quantum property regressions for 12 targets on QM9.}\label{tab:qm9_joint}
\begin{tabular}{l|ccccccccc|c}
\toprule
Method & GCN     & RGCN    & ChebyNet & GAT     & GGNN    & LanczosNet & GIN & sMPNN   & MPNN   & IGNN            \\
\midrule
nMAE             & 0.1404 & 0.1024 & 0.1325   & 0.1297 & 0.1003 & 0.1011  & 0.0999 & 0.0892 & 0.0400 & \textbf{0.0352} \\
MAE              & 5.5665 & 3.8075 & 4.9839   & 5.1818 & 3.7037 & 3.7379  & 3.4623 & 3.1561 & 0.7162 & \textbf{0.6083} \\
\midrule
mu               & 0.5695 & 0.5088 & 0.5365   & 0.5582 & 0.5261 & 0.4716  & 0.4841 & 0.4722 & 0.1076 & \textbf{0.0999} \\
alpha            & 0.9216 & 0.6421 & 0.9032   & 0.8482 & 0.6192 & 0.6486  & 0.6239 & 0.5171 & 0.3336 & \textbf{0.2896} \\
HOMO $(10^{-3})$ & 5.4527 & 4.4305 & 5.0602   & 5.1977 & 4.5269 & 3.9553  & 4 2102 & 3.7973 & 2.4983 & \textbf{2.1932} \\
LUMO $(10^{-3})$ & 6.4143 & 5.2910 & 6.0505   & 5.9956 & 5.2528 & 4.2660  & 4.7796 & 4.6577 & 2.8437 & \textbf{2.5304} \\
gap $(10^{-3})$  & 8.1872 & 6.5703 & 7.6031   & 7.7851 & 6.6978 & 6.4862  & 6.1536 & 5.7094 & 3.6178 & \textbf{3.2005} \\
R2               & 54.793 & 40.076 & 48.779   & 52.257 & 40.354 & 36.013  & 34.754 & 33.348 & 6.2450 & \textbf{5.3776} \\
ZPVE $(10^{-3})$ & 2.6059 & 1.4666 & 3.0258   & 2.0900 & 1.3166 & 1.4557  & 1.7297 & 1.3708 & 0.6778 & \textbf{0.6199} \\
U0               & 2.4986 & 1.0324 & 2.2690   & 2.0100 & 0.6517 & 1.8753  & 1.3400 & 0.8142 & 0.4408 & \textbf{0.3514} \\
U                & 2.4986 & 1.0324 & 2.2690   & 2.0100 & 0.6502 & 1.8191  & 1.3400 & 0.8142 & 0.4407 & \textbf{0.3514} \\
H                & 2.4986 & 1.0324 & 2.2690   & 2.0100 & 0.6535 & 1.8495  & 1.3400 & 0.8142 & 0.4407 & \textbf{0.3514} \\
G                & 2.4986 & 1.0324 & 2.2691   & 2.0101 & 0.6497 & 1.8383  & 1.3401 & 0.8143 & 0.4407 & \textbf{0.3514} \\
Cv               & 0.4970 & 0.3154 & 0.4905   & 0.4558 & 0.3220 & 0.3230  & 0.3077 & 0.2627 & 0.1356 & \textbf{0.1183} \\
\bottomrule
\end{tabular}
\end{threeparttable}
\end{table}

\subsubsection{Joint learning}

We first study the setting of joint learning on all 12 targets simultaneously. We report the normalized MAE (nMAE; averaged over normalized target properties), MAE (averaged over target properties in their original scale), and the individual MAE for each target in their original scale, in Table \ref{tab:qm9_joint}.

Experimental results demonstrate that our proposed IGNN outperforms existing baselines on all 12 targets by a significant margin. GCN, ChebyNet, and GAT focus on learning node representation, with edge information like bond types and distances largely ignored. RGCN, GGNN, and LanczosNet incorporate bond types into the model by learning a separate set of parameters for a specific type of bond, and thus have better performances (nMAEs and MAEs are about 0.10 and 3.7, respectively). However, they cannot deal with continuous edge features.

Continuous edge features such as distances between atoms are closely related to quantum properties. For example, the smaller the distance between the two atoms, the stronger the bond is, and consequently a higher bond energy is associated with this atom pair. MPNN improves previous models by proposing an edge network, allowing inputs of vector valued edge features. Compared sMPNN to MPNN, we further validate the importance of distances in the molecular graphs. 
By maximizing the mutual information of edge features and the transform matrix, our proposed IGNN achieves the lowest error, decreasing by {\bf 12\%} in nMAE and {\bf 15\%} in MAE compared to the second best model, MPNN.

\begin{table}[t]
\centering
\scriptsize
\begin{threeparttable}
\caption{Regression results for R2 on QM9.}\label{tab:qm9_separate}
\begin{tabular}{c|ccccccccc|c}
\toprule
Method & GCN      & RGCN      & ChebyNet & GAT      & GGNN     & LanczosNet   & GIN    & sMPNN & MPNN    & IGNN             \\
\midrule
MAE             & 40.9216  & 24.3837  & 28.9985  & 34.0738  & 25.5890  & 26.5068    & 22.7349 & 22.9573  & 1.2002  & \textbf{0.8595}  \\
\bottomrule
\end{tabular}
\end{threeparttable}
\end{table}

\subsubsection{Learning properties separately}
Next, we train one model for each individual target, and select the property R2 for demonstration since R2 seems to be the most challenging task in the previous experiments. Experimental results are shown in Table \ref{tab:qm9_separate}. We found that for all models, training a sperate model per target has a superior performance than jointly training on all 12 targets, which is consistent with the observation in \cite{gilmer2017neural}. IGNN substantially outperforms other state-of-the-art baselines.

\begin{table}[t]
\centering
\scriptsize
\begin{threeparttable}
\caption{Generalization capability from QM9 to larger molecular graphs. Models are fine tuned on 1000 larger molecules of our own dataset.
}\label{tab:transfer}
\begin{tabular}{l|ccccccccc|c}
\toprule
Method           & GCN      & RGCN     & ChebyNet & GAT      & GGNN     & LanczosNet & GIN     & sMPNN    & MPNN    & IGNN   \\
\midrule
nMAE             & 0.2444   & 0.2351   & 0.2706   & 0.2387   & 0.2268   & 0.1887          & 0.1901          & 0.2019          & \textbf{0.1825}  & 0.1860          \\
MAE              & 16.662   & 13.980   & 17.084   & 15.717   & 14.486   & 10.612          & 11.027          & 11.351          & 5.3396           & \textbf{5.0229} \\
\midrule
mu               & 0.9538   & 0.9521   & 0.9995   & 0.9421   & 1.0000   & \textbf{0.7719} & 0.7945          & 0.9554          & 0.8039           & 0.8007          \\
alpha            & 1.9321   & 1.9090   & 2.2516   & 2.0125   & 1.7081   & 1.5038          & 1.6473          & \textbf{1.4864} & 1.5252           & 1.6968          \\
HOMO $(10^{-2})$ & 1.0324   & 1.1512   & 1.1588   & 1.0244   & 1.0033   & 0.9049          & \textbf{0.8727} & 1.0040          & 1.0774           & 1.1399          \\
LUMO $(10^{-2})$ & 1.1829   & 1.2819   & 1.3766   & 1.1027   & 1.2317   & \textbf{1.0042} & 1.0212          & 1.1078          & 1.2546           & 1.2611          \\
gap $(10^{-2})$  & 1.6022   & 1.7861   & 1.7880   & 1.5540   & 1.6073   & 1.4115          & \textbf{1.3972} & 1.5423          & 1.6965           & 1.7111          \\
R2               & 151.61   & 120.76   & 137.36   & 135.77   & 132.59   & 95.431          & 98.487          & 104.92          & \textbf{31.074}  & 31.407          \\
ZPVE $(10^{-3})$ & 4.2120   & 2.7141   & 5.2599   & 3.9581   & 2.4051   & \textbf{2.2108} & 3.0573          & 2.2797          & 2.7610           & 3.3199          \\
U0               & 11.122   & 10.835   & 15.848   & 12.239   & 9.4366   & 7.3919          & 7.6757          & 7.0417          & 7.3896           & \textbf{6.4027} \\
U                & 11.122   & 10.835   & 15.848   & 12.239   & 9.4312   & 7.1778          & 7.6728          & 7.0417          & 7.3853           & \textbf{6.4007} \\
H                & 11.122   & 10.835   & 15.848   & 12.239   & 9.4419   & 7.1148          & 7.6759          & 7.0418          & 7.3880           & \textbf{6.4051} \\
G                & 11.122   & 10.834   & 15.848   & 12.239   & 9.4266   & 7.2728          & 7.6741          & 7.0412          & 7.3853           & \textbf{6.4014} \\
Cv               & 0.9118   & 0.7516   & 0.9570   & 0.8779   & 0.7566   & \textbf{0.6406} & 0.6553          & 0.6447          & 0.6770           & 0.7166          \\
\bottomrule
\end{tabular}
\end{threeparttable}
\end{table}

\subsubsection{Generalization on larger molecular graphs}

In practical applications of material science and drug discovery, researchers are often interested in generalizing the knowledge learned from small-size molecules to large-size molecules. While most GNNs are designed to address homogeneous datasets, which contains similar-size molecules, the development of models with excellent generalization capability is of great importance, especially for the scenarios when it is expensive to obtain sufficient data for large-size molecules.

In this experiment, we aim to test the generalization capability of the models on larger and more diverse graphs. The results are shown in Table \ref{tab:transfer}. We use the model trained in Sec. \ref{subsect:quantum_chemistry} on QM9 as an initialization, and select 1000 more molecules from a proprietary dataset\footnote{The dataset will be released soon.} to train 300 epochs. Performance is evaluated on a test set of 10,000 molecules. Our proprietary dataset contains molecules with 10 heavy atoms. Different from QM9, the target properties are of higher precision and the molecules are larger and more diverse. Note that this task is challenging since only a small subset (1000 molecules) of the data can be seen by the models, and the distribution of the target properties is different from QM9.

We observe that LanczosNet, GIN, sMPNN, MPNN, and IGNN can achieve satisfying results. However, only MPNN and IGNN perform well on the property R2. It is because MPNN and IGNN is good at exploiting the distance information on edges, whereas others cannot. Especially, IGNN outperforms other baselines on  MAE and 1/3 of the tasks since maximizing mutual information in IGNN can be interpreted as regularization on edges, and thereby better generalization ability can be achieved.

\begin{table}[t]
\centering
\scriptsize
\begin{threeparttable}
\caption{pIC$_{50}$ prediction on ChEMBL}\label{tab:chembl}
\begin{tabular}{c|ccccccccc|c}
\toprule
Method   & GCN    & RGCN   & ChebyNet & GAT    & GGNN   & LanczosNet & GIN    & sMPNN  & MPNN   & IGNN            \\
\midrule
Test MAE & 0.7053 & 0.6702 & 0.6408   & 0.6465 & 0.6727 & 0.6317 & 0.6226 & 0.6589 & 0.6435 & \textbf{0.6174} \\
Test R   & 0.7946 & 0.8091 & 0.8153   & 0.8272 & 0.8085 & 0.8196 & 0.8246 & 0.8133 & 0.8206 & \textbf{0.8350} \\
\bottomrule
\end{tabular}
\end{threeparttable}
\end{table}

\subsection{Drug discovery}
\begin{figure}
	\centering{\includegraphics[width=0.45\columnwidth]{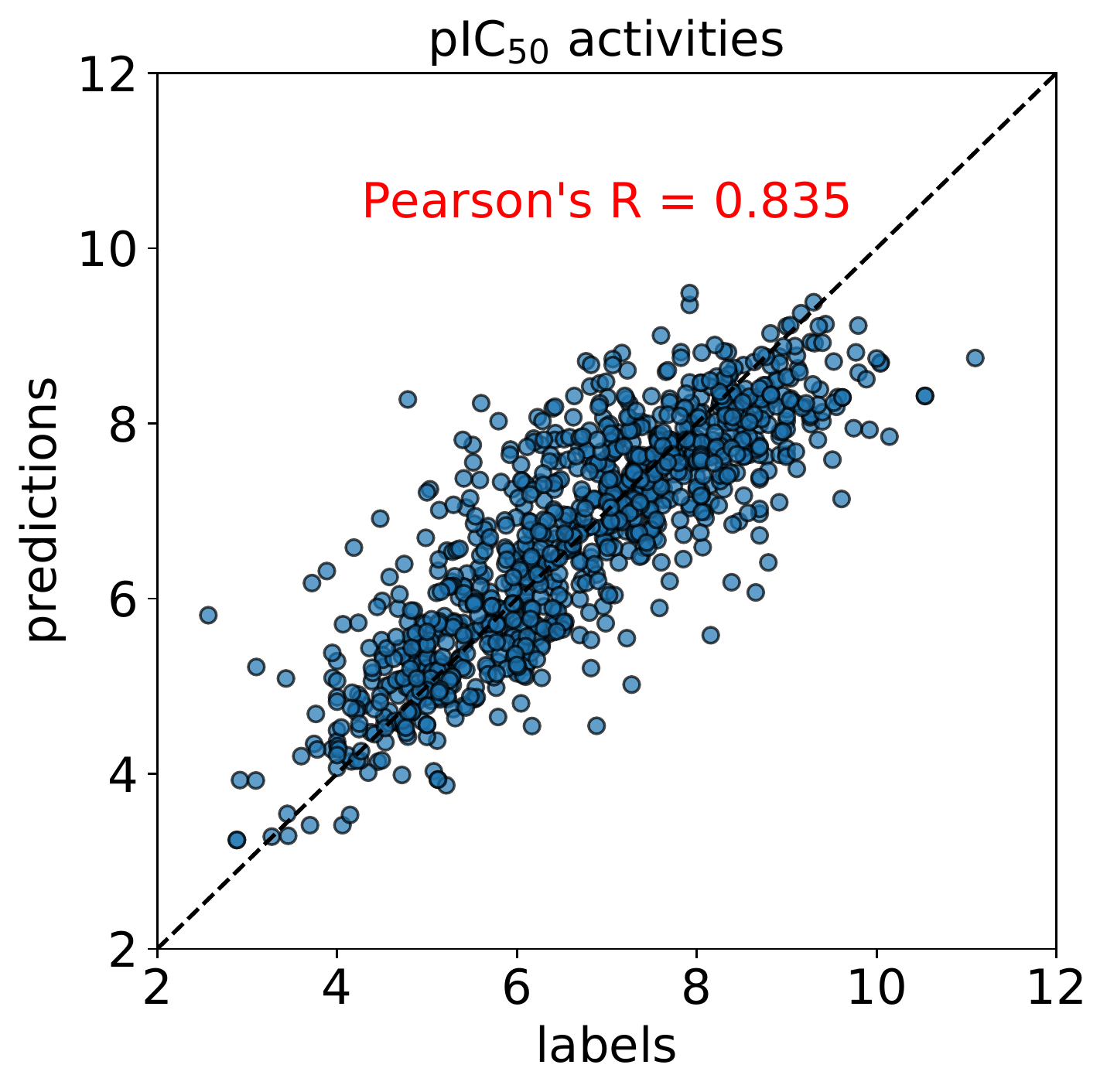}}
	\vskip -0.15in
	\caption{Correlation analysis between true and predicted pIC$_{50}$.}\label{fig:chembl_correlation}
	\vskip -0.2in
\end{figure}

Epidermal Growth Factor Receptor (EGFR; ErbB-1) is a protein target found on the surface of some cells and to which epidermal growth factor binds, causing the cells to divide excessively. ErbB-1 are found in many human cancers, and their over-expression or constitutive activation may be critical factors in the development and malignancy of these tumors. Inhibition of ErbB pathways with targeted agents has been shown to be validated anti-cancer therapy.

In order to discover new inhibitors to the target ErbB-1, we use pIC$_{50}$ to measure the potency of a molecule in inhibiting ErbB-1. Accurate prediction of the pIC$_{50}$ can significantly accelerate the drug design process, as no in vitro determination of pIC$_{50}$ is needed.  We extracted 7,698 molecules with pIC$_{50}$ value from the ChEMBL database~\cite{gaulton2011chembl} in total. The retained dataset contains molecules with up to 100 heavy atoms. We randomly select 1000 samples for validation, 1000 samples for testing, and the remaining for training. We follow the same setting as in Sec. \ref{subsect:quantum_chemistry} except the number of the target property is one. Besides MAE, we also report the Pearson's correlation coefficient ($R\in[-1, 1]$) between the true and predicted pIC$_{50}$ in Table \ref{tab:chembl}. A higher R represents the predicted value being more positively correlated with the true value.


Molecular graphs in real-world drug discovery task are generally large and of high diversity.  From Table \ref{tab:chembl}, we observe that our proposed IGNN is able to predict pIC$_{50}$ with $R=0.835$ and $\text{MAE}=0.6174$, achieving the best perfomrance. The illustration of the correlation analysis for IGNN is depicted in Figure \ref{fig:chembl_correlation}. Our experimental study shows that IGNN works well in such complex setting, verifying the importance of information maximization on edges.

%% file: sec_Conclusions.tex
In this work, we designed an Information Maximizing Graph Neural Networks (IGNN) to maximize the mutual information between edge feature vectors and transform matrices. Currently, no other GNN variants utilize the edge information as efficiently as IGNN. The key element critical to IGNN's success is the reformulation of the mutual information as a differentiable objective by adopting a variational approach. We tested IGNN in various scenarios including quantum property regressions and drug discovery. The promising results demonstrated the effectiveness of IGNN. 